\documentclass[]{spie}  

\usepackage{amsmath,amsfonts,amssymb}
\usepackage{graphicx}
\usepackage{siunitx}
\usepackage{booktabs}
\usepackage[noend]{algpseudocode}
\usepackage{algorithm}
\usepackage{multirow}
\usepackage[colorlinks=true, allcolors=blue]{hyperref}

\title{Two-dimensional Deep Regression for\\Early Yield Prediction of Winter Wheat}

\author{Giorgio Morales}
\author{John W. Sheppard}
\affil{Gianforte School of Computing, Montana State University, Bozeman, MT, 59717 USA}

\authorinfo{Further author information: (Send correspondence to J.W.S.)\\G.M.: E-mail: giorgiol.moralesluna@student.montana.edu\\  J.W.S.: E-mail: john.sheppard@montana.edu}

\begin{document} 
\maketitle

\begin{abstract}
Crop yield prediction is one of the tasks of Precision Agriculture that can be automated based on multi-source periodic observations of the fields. 
We tackle the yield prediction problem using a Convolutional Neural Network (CNN) trained on data that combines radar satellite imagery and on-ground information. 
We present a CNN architecture called Hyper3DNetReg that takes in a multi-channel input image and outputs a two-dimensional raster, where each pixel represents the predicted yield value of the corresponding input pixel. 
We utilize radar data acquired from the Sentinel-1 satellites, while the on-ground data correspond to a set of six raster features: nitrogen rate applied, precipitation, slope, elevation, topographic position index (TPI), and aspect. 
We use data collected during the early stage of the winter wheat growing season (March) to predict yield values during the harvest season (August).
We present experiments over four fields of winter wheat and show that our proposed methodology yields better results than five compared methods, including multiple linear regression, an ensemble of feedforward networks using AdaBoost, a stacked autoencoder, and two other CNN architectures. 
\end{abstract}


\section{INTRODUCTION}
\label{sec:intro}  

Precision Agriculture (PA) is a crop management technique that uses Information Technologies (ITs) to obtain site-specific information about the condition of the crops. 
This knowledge is then used to elaborate strategies to make efficient use of the available farming resources (e.g., water, nutrients, and pesticides), avoiding waste, minimizing environmental impact, and maximizing profit~\cite{future,PAapplications,PAreview}. To achieve this, models are developed that relate input factors, such as environmental and management variables, and outcome variables, such as the crop yield. These models can be obtained through On-Farm Precision Experimentation (OFPE), which generates site-specific data on responses to field management\cite{bullock}.

Recently, PA has benefited from the confluence of the growing availability of sensors that can accurately and continuously collect information about fields~\cite{IOUT,UAS}, the boom of machine learning, and the development of accessible and fast computational resources~\cite{yieldML,CVreview}. 
In this context, one of the most benefited areas of PA is crop yield prediction, which is important for national food security as it allows to prepare import/export policies and to estimate prices~\cite{HORIE}. 
Crop yield prediction also provides the farmers with tools to make informed decisions, such as designing accurate marketing plans for their products~\cite{HORIE} or determining the nitrogen fertilizer rates needed in specific regions of the field to maximize farmer profits~\cite{maxwell,bullock}.

Many machine learning approaches have been proposed lately to automate the yield prediction task. 
Some of them rely only on remotely sensed data, such as Moderate Resolution Imaging Spectroradiometer (MODIS) or Sentinel satellite imagery~\cite{msthesis,potato}, while others incorporate on-ground data, such as soil electroconductivity or nitrogen rate \cite{barbosa,peerlinck2019adaboost}. 
The common goal is to train a regression model to estimate the crop yield in terms of bushels per acre (bu/ac) as accurately as possible given some input information. 
All previous works have focused on predicting the yield values of single georeferenced points of the field, which represent small regions of the field (e.g. a point may represent an area of $10 \times 10$ \SI{}{\meter}). 
Hence, the regression models proposed by previous works process the information of each point of the field and (possibly) from its surroundings to estimate its corresponding yield value.

Assuming that the yield value of a point of a field is not independent of those of its surroundings, it would be natural to consider a yield prediction model that analyzes spatial neighborhoods of points and predicts the yield values of all the points within these neighborhoods. 
Therefore, in this paper, we present a convolutional neural network for yield prediction called Hyper3DNetReg. 
It takes as an input a multi-channel image that corresponds to a small neighborhood of points of the field and each channel represents a different feature of that neighborhood (e.g., nitrogen rate applied). 
Given a two-dimensional input, our Hyper3DNetReg network outputs a two-dimensional yield raster, where each pixel represents the predicted yield value of its corresponding input pixel.
Our models are field-specific, which means they are trained on data of a given field from previous years (excluding the last observed year) and used to predict yield maps using data from the last observed year of the same field. 
Finally, we hypothesize that our Hyper3DNetReg network with a two-dimensional output will lead us to generate more accurate predicted yield maps than the compared methods with a single output.

The remainder of this paper is structured as follows: In Section~\ref{sec:related}, we provide a brief review of previous related work done with machine learning and crop yield prediction. Section~\ref{sec:method} provides further details about the datasets used in this work, as well as how we pre-processed the data for our experiments. In this section, we also describe our Hyper3DNetReg network architecture and how it is used to generate predicted yield maps. Section \ref{sec:results} presents the results of our experiments, and  Section~\ref{sec:discussion} discusses those results. Finally, Section \ref{sec:conclusions} offers concluding remarks.

\section{Related Work} \label{sec:related}

Crop yield prediction is one of the most studied tasks of PA. 
Classic approaches involve the use of simple linear and non-linear regression models based on remotely sensed vegetation indexes such as Normalized Difference Vegetation Index (NDVI) or Normalized Difference Water Index (NDWI)~\cite{reg1,reg2}. 
Other approaches combine on-ground data and remotely sensed data using more advanced machine learning techniques like support vector machines, random forests, regression trees, or $k$-nearest neighbors~\cite{svm,rf,regtree}.

Moreover, many approaches based on artificial neural networks have been shown to outperform classic methods in the context of yield prediction. 
For instance, Peerlinck \textit{et al.} \cite{Peerlinck2018UsingDL} used a deep neural network based on a stacked autoencoder and spatially sampled data consisting of factors such as previous years' nitrogen application rate, field slope, NDVI, and precipitation. 
The results showed improvements over other shallow neural networks and simple linear and non-linear regression models. Peerlinck \textit{et al.} \cite{peerlinck2019adaboost} also proposed to use an Approximation AdaBoost algorithm with feedforward neural networks (FNNs) as weak learners. This method approximates the loss function by using a threshold approach that discards small errors during the weight update of the weak learners. Experimental results showed that this approach  gives better results than using a simple FNN.

CNNs are a way to explicitly take advantage of the spatial information of small regions of the field. 
For that reason, You \textit{et al.} \cite{deepgauss} proposed the use of CNNs and Long-Short Term Memory (LSTM) networks for yield prediction, along with a Gaussian process component to model the spatio-temporal structure of the data. 
Similarly, Russello \textit{et al.}~\cite{msthesis} proposed a 3-D CNN architecture for spatio-temporal feature learning achieving significant improvements over the results obtained by You \textit{et al.}~\cite{deepgauss}. 
Barbosa \textit{et al.}~\cite{barbosa} proposed to use small two-dimensional CNNs that take as inputs a set of five raster features: nitrogen rate, seed rate, elevation map, and soil’s shallow electroconductivity. 
Results showed a substantial improvement over those obtained by multiple linear regression and random forest. 
It is worth noting that the output of all of the previously mentioned studies is a single predicted yield value corresponding to one specific point or cell of the studied region. 
That is, even in the case of CNNs where the input consists of a small neighborhood of points, the output of the network corresponds to the predicted yield value of the central pixel of the input neighborhood. 

In this work, we explore the use of Synthetic Aperture Radar (SAR) images captured by the Sentinel-1 mission for predicting yield. 
They represent an advantage over optical satellite images because the emitted microwaves penetrate through clouds so they are insensitive to atmospheric conditions. 
Backscattering coefficients of SAR images provide information about soil conditions such as roughness or presence of vegetation, which can be used for soil moisture retrieval~\cite{moisture1,moisture2} or detection of change in vegetation~\cite{vegetation}. 
Previous works used series of Sentinel-1 images corresponding to the entire growing season to estimate the production of winter wheat, rapeseed, and rice crops~\cite{biomass,rice}. 
Furthermore, Zhuo \textit{et al.}~\cite{moistureyield} used Sentinel-1 and Sentinel-2 images to retrieve soil moisture and incorporate this information into the World Food Studies (WOFOST) model~\cite{van1989wofost} in order to improve the yield estimation accuracy of winter wheat.

\section{Proposed Methodology}  \label{sec:method}

\subsection{Datasets} 

\begin{figure}[t!]
\centering
\includegraphics[height = 7.5cm]{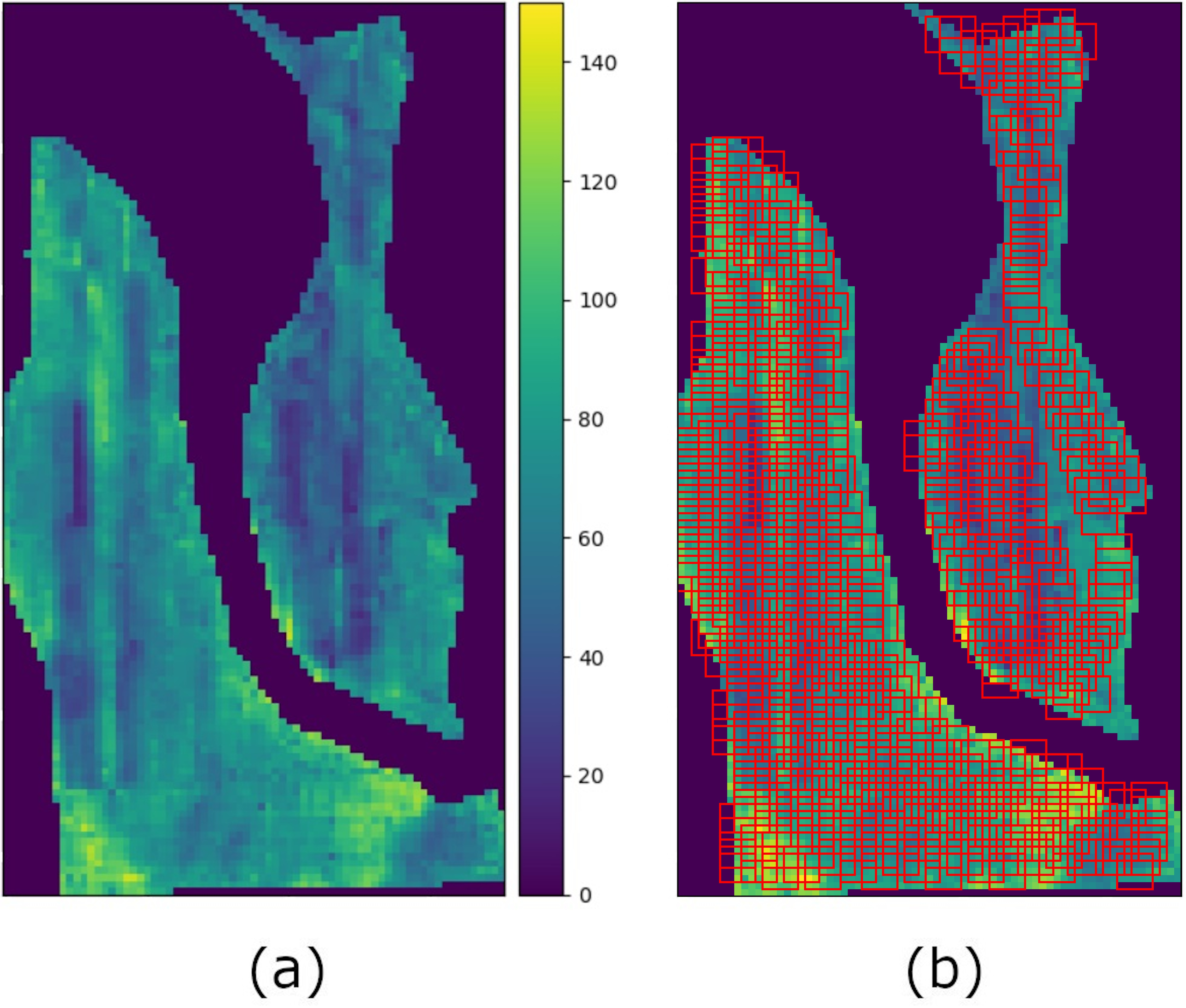}
\caption{Image rasters corresponding to the Sec35mid field. \textbf{(a)} Yield raster. \textbf{(b)} Automatic extraction of $5 \times 5$ pixel patches.}
\label{fig:dataset}
\end{figure}

Four fields of winter wheat from two different farms from Montana were selected for our experiments.
The yield maps correspond to the harvest season (i.e. August for all the studied fields) and are acquired using a yield monitor mounted on a combine harvester so that we obtain georeferenced yield values --measured in bushels per acre (bu/ac)-- every few feet while traveling through the field.
The acquired data is then aggregated at a scale of \SI{10}{\meter} to obtain a yield map with equally spaced points. 
To do this, the whole field is divided in a grid where each cell represents a region of $10 \times 10$~\SI{}{\meter}.
Then if more than one point is found within a cell, we calculate their mean yield value to represent the yield value of that cell. 
The resulting yield map can be interpreted as an image raster such as that shown in Figure~\ref{fig:dataset}.a.

While the yield value represents the response variable in our regression problem, the explanatory variables correspond to the SAR satellite images acquired by the Sentinel-1 mission and the on-ground data. 
Sentinel-1 images contain two bands acquired using Vertical Transmit-Vertical Receive Polarisation (VV) and Vertical Transmit-Horizontal Receive Polarisation (VH). 
These images were obtained at the Ground Range Detected (GRD) level, which includes three pre-processing steps: noise removal, radiometric calibration, and ortho-rectification. 
The resulting images have a spatial resolution of \SI{10}{\meter}. On the other hand, the on-ground data corresponds to a set of six raster features: nitrogen rate applied, precipitation, slope, elevation, topographic position index (TPI), and aspect. 
Thus, our resulting input can be seen as an image datacube with eight channels in total where each pixel has a resolution of $10 \times 10$~\SI{}{\meter} (i.e., the same resolution of our yield maps). 
It is of particular importance to note that the nitrogen fertilizer is applied in March and the acquired Sentinel-1 images correspond to the same month. 
In that sense, we use data acquired in March to predict crop yield values in August of the same year.

Given that each field is represented with one large raster image, we have to divide it into small patches. 
Thus, we automatically extract square patches using a $5\times 5$ pixel window allowing a maximum overlapping of 0.75 so that we collect a sufficient number of training samples from one single field, as shown in Figure~\ref{fig:dataset}.b. 
We use the information collected from previous years to construct our training and validation datasets (we use 90\% of the data for training and 10\% for validation). Furthermore, we use the information from the last observed year to test the prediction error of our models.
Table~\ref{tab:fields} shows the total number of training and validation samples and the years of observation for each field used in our experiments. 

\begin{table}[t]
\caption{Number of samples and years of observation for each field.}
\label{tab:fields}
\begin{center}
    \begin{tabular}{|c|c|c|c|}
\hline
\textbf{Field} & \textbf{\begin{tabular}[c]{@{}c@{}}Training + Validation \\ Samples\end{tabular}} & \textbf{Training Years} & \textbf{Testing Year} \\ \hline
Henrys & 690 & 2016, 2018 & 2020 \\ \hline
Sec1west &  1812    & 2016, 2019 & 2021 \\ \hline
Sec35mid & 960 & 2016, 2018 & 2020 \\ \hline
Sec35west & 265 & 2017 & 2020 \\ \hline

\end{tabular}
\end{center}
\vspace{2ex}
\end{table}

\subsection{Yield Prediction Model} \label{prediction}

Let $X$ be a two-dimensional  multi-channel input image of $W \times W$ pixels with $n$ channels. 
For the datasets used in this work, we consider $W=5$ and $n=8$. 
Furthermore, let $Y$ be a two-dimensional output image of $N \times N$ pixels ($N \leq W$) where the value of its $(i, j)$--th pixel represents the yield value corresponding to the $(i + \frac{W - N}{2}, j + \frac{W - N}{2})$--th pixel of $X$.

We construct models that capture the association between the output $Y$ and the input $X$ using a CNN called Hyper3DNetReg as a regression model.
Let $f(\cdot)$ denote the function computed by our Hyper3DNetReg network and $\theta$ denote the weights of the network. Thus, the predicted yield $\hat{Y}$ given input $X$ is calculated as:

\begin{equation*}
    \hat{Y} = f(X).
\end{equation*}

\begin{figure}[!t]
\centering
\includegraphics[width = \textwidth]{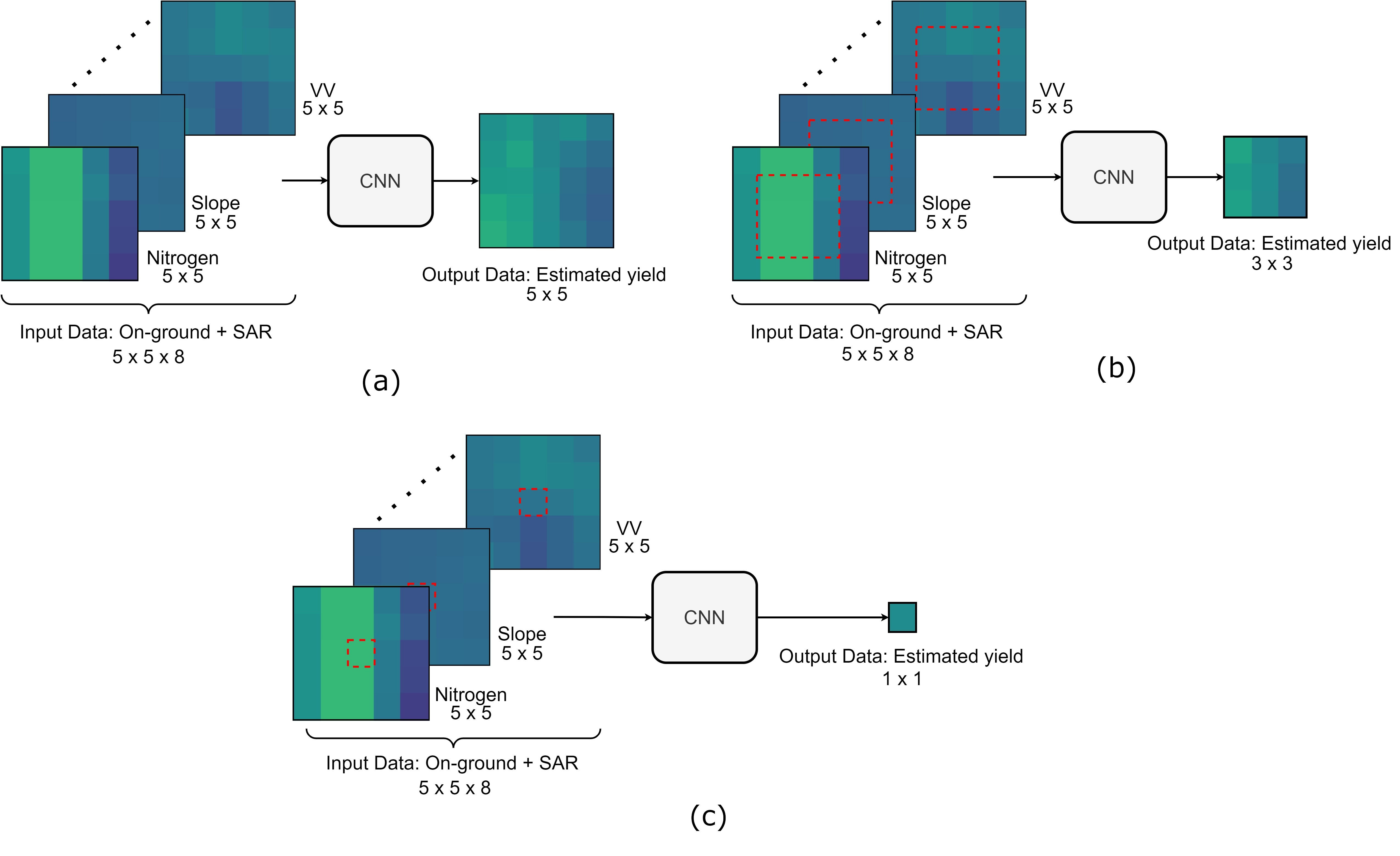}
\caption{Proposed yield prediction model using different output window sizes: \textbf{(a)} $5\times 5$, \textbf{(b)} $3\times 3$, and \textbf{(c)} $1\times 1$. } 
\label{fig:model}
\end{figure}

\noindent We implement models using three output window sizes (i.e., $N = 5, \; 3, \; 1$) to evaluate the impact of $N$ on the effectiveness of our method. 
Fig.~\ref{fig:model} depicts our proposed yield prediction models. 
The output windows shown in Fig.~\ref{fig:model}.b and Fig.~\ref{fig:model}.c represent the predicted yield of the area enclosed by the dotted squares within the input images.

\setlength{\tabcolsep}{4pt}
\begin{table}
\caption{Hyper3DNetReg architecture.}
\label{tab:net}
\begin{center}

\begin{tabular}{|c|c|c|c|}
\hline
\textbf{Layer Name}   & \textbf{Kernel Size} & \textbf{Padding Size} & \textbf{Output Size} \\ \hline
Input                 & ---                & ---              & (5, 5, $n$, 1) \\ \hline
Conv3D + ReLU + BN         & (3, 3, 3)          & (1, 1, 1)      & (5, 5, $n$, 32) \\\hline
Conv3D + ReLU + BN         & (3, 3, 3)          & (1, 1, 1)      & (5, 5, $n$, 32) \\ \hline
CONCAT                & ---          & ---      & (5, 5, $n$, 64) \\ \hline
Conv3D + ReLU + BN         & (3, 3, 3)          & (1, 1, 1)      & (5, 5, $n$, 32) \\ \hline
CONCAT                & ---          & ---      & (5, 5, $n$, 96) \\ \hline
Conv3D + ReLU + BN         & (3, 3, 3)          & (1, 1, 1)      & (5, 5, $n$, 32) \\ \hline
CONCAT                & ---          & ---      & (5, 5, $n$, 128) \\ \hline
Reshape               & ---       & ---     & (5, 5, $128 \cdot n$)     \\ \hline
Dropout (0.5)         & ---       & ---     & (5, 5, $128 \cdot n$)     \\ \hline
SepConv2D + ReLU + BN & (3, 3)     & (1, 1)   & (5, 5, 512)        \\ \hline
SepConv2D + ReLU + BN & (3, 3)     & (1, 1)   & (5, 5, 320)        \\ \hline
Dropout (0.5)         & ---       & ---     & (5, 5, 320)      \\ \hline
SepConv2D + ReLU + BN & (3, 3)     & (1, 1)   & (5, 5, 256)        \\ \hline
Dropout (0.5)         & ---       & ---     & (5, 5, 256)      \\ \hline
SepConv2D + ReLU + BN & (3, 3)     & (1, 1)   & (5, 5, 128)        \\ \hline
SepConv2D + ReLU + BN & (3, 3)     & (1, 1)  & (5, 5, 32)        \\ \specialrule{1.5pt}{1pt}{1pt}
\multicolumn{4}{|l|}{ \textbf{if} $N = 5$ or $N = 3$:} \\ \hline
Conv2D + ReLU         & (3, 3)     & $\begin{cases}
    (1, 1), & \text{if } N = 5\\
    (0, 0),              & \text{elif } N = 3
\end{cases}$   & \textbf{($N$, $N$, 1)}        \\ \specialrule{1.5pt}{1pt}{1pt}

\multicolumn{4}{|l|}{ \textbf{elif} $N = 1$:} \\ \hline
Conv2D + ReLU         & (3, 3)     & (0, 0)  & (3, 3, 1)  \\ \hline 
Reshape               & ---       & ---     & (9, 1)     \\ \hline
FC               & ---       & ---     & \textbf{$N$}     \\ \hline
\end{tabular}

\end{center}
\end{table}

\subsection{Hyper3DNetReg Architecture}

The architecture of the CNN that is used in this work is based on that of the Hyper3DNet network~\cite{hyper3dnet}, which is a 3D--2D CNN designed for hyperspectral image classification. 
This architecture is suitable for our task because, similar to the case of hyperspectral images, we are dealing with two-dimensional inputs with multiple channels; therefore, the 3-D convolutional filters of the first module of the Hyper3DNet network would allow us to better capture the interaction between the input channels and not only the spatial neighborhoods.

Nevertheless, two key differences prevent us from using the original Hyper3DNet architecture. 
The first one is that Hyper3DNet is designed for classification tasks while ours is a regression task. 
The second difference is that the output of the original Hyper3DNet is a one-dimensional vector containing the scores of each of the possible classes while our output is a two-dimensional patch with dimensions $N \times N$ (where $N =$ 5 or 3) or a single real value ($N=1$). 
Thus, we need to modify the Hyper3DNet architecture to fit our needs.

The modified network is referred to as Hyper3DNetReg and its architecture is shown in Table~\ref{tab:net}, where $n$ denotes the number of channels in the input. 
``Conv3D" denotes a 3-D convolutional layer, ``Conv2D" denotes a 2-D convolutional layer, ``SepConv2D" denotes a 2-D separable convolutional layer, ``FC" denotes a  fully-connected layer, and ``ReLU" denotes a rectified linear unit activation layer (where $ReLU(x)=\max (0, x))$. 
Furthermore, the ``CONCAT" layer concatenates the outputs of the two preceding convolutional blocks along the fourth dimension.  
We also included three ``Dropout" layers which randomly zero out some elements of the preceding tensor with a probability of 0.5, but only during training.
Depending on whether the output is a two-dimensional patch or a single value, the last layer of the network is a 2-D convolutional layer or a fully-connected layer, respectively.

In addition, we modified the second module of the Hyper3DNet network (i.e., the 2-D spatial encoder) to use a stride and padding of $(1,1)$ so that the spatial dimensions remain the same throughout the network.
This is because if we use a $5 \times 5$ input ($W=5$) and desire to obtain an output of the same dimensions ($N=5$), the feature maps obtained by the inner blocks of the network should maintain their spatial dimensions as well. 
Therefore, the inner feature maps of the network must not be affected by any downsampling. 
We also removed the final fully connected layer of the original Hyper3DNet network, and we replaced it with a convolutional block that reduces the number of channels to one. 
Hence, we avoid using an excessive number of parameters.
For instance, if the size of the output window was set to $N=5$, the last convolutional block would require only 289 trainable parameters.
In contrast, if we used a fully-connected layer with 25 outputs in place of the last convolutional block, this layer would require 20,025 trainable parameters. 
However, for the case that the output of the model be a single value ($N=1$), it is safe to add a fully-connected layer with one output to the last convolutional block, as it only adds nine trainable parameters.

\subsection{Predicted Yield Map Generation} \label{generate}

\begin{figure}[t!]
\centering
\includegraphics[width = 15cm]{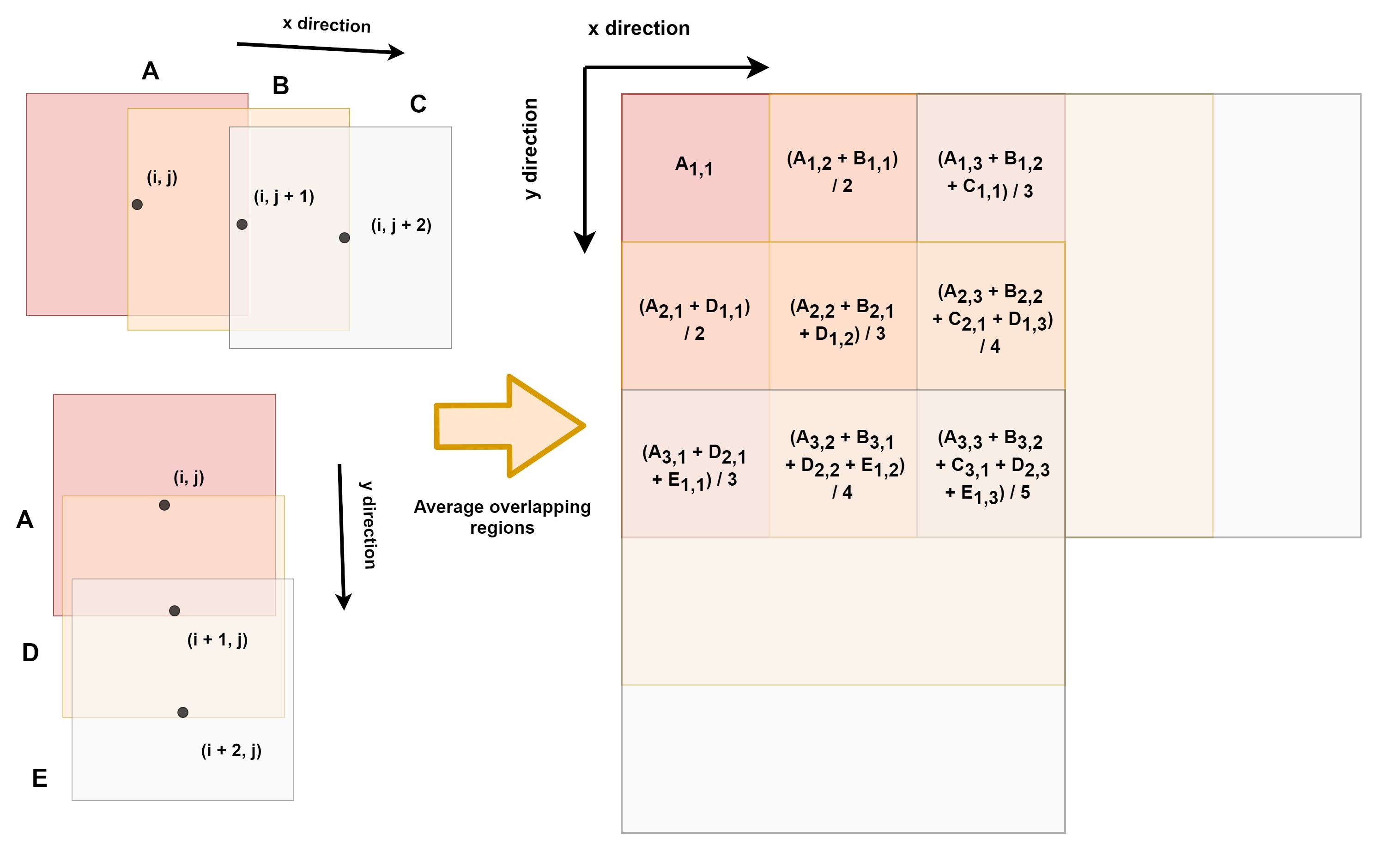}
\caption{Yield map generation using overlapping patches.  } 
\label{fig:overlap}
\vspace{2ex}
\end{figure}

Our goal is to estimate the yield of the entire field. 
To do this, we create a two-dimensional $5 \times 5$ patch around each point of the field and process it through our Hyper3DNetReg network to estimate the yield of an $N \times N$ patch around that point, as explained in Sec.~\ref{prediction}. 
Using this approach, the resulting predicted yield patches of consecutive points will overlap. Fig.~\ref{fig:overlap} illustrates a $3 \times 3$ yield patch \textbf{A} ($N=3$) generated around a point with coordinates $(i, j)$. 
For the sake of brevity, in this example, the $(i, j)$ point has only four point neighbors: $(i, j + 1)$, $(i, j + 2)$, $(i + 1, j)$, and $(i + 2, j)$, which generate the yield patches \textbf{B}, \textbf{C}, \textbf{D}, and \textbf{E}, respectively. 
In reality, however, we utilize all available neighbors to calculate the average.
Hence, we show that the pixel values of the overlapping regions are averaged to obtain the final yield estimation map. 
By doing so, we are effectively smoothing the yield estimations from different patches; thus, avoiding noisy results and reducing uncertainty.

\section{Results} \label{sec:results}

In this section, we test our yield map prediction methodology using the Hyper3DNetReg network and five other yield prediction methods: Approximate AdaBoost (AdaBoost.App)~\cite{peerlinck2019adaboost}, a stacked autoencoder (SAE)~\cite{Peerlinck2018UsingDL}, a three-dimensional CNN (3D-CNN)~\cite{msthesis}, late fusion CNN (CNN-LF)~\cite{barbosa}, and multiple linear regression (MLR). 
For AdaBoost.App, we use ten feedforward neural networks as weak learners. 
Each network consists of three hidden layers, where 100, 96, and 32 hidden nodes were used from the top to bottom layer respectively.
For SAE, we use a feedforward neural network with three hidden layers, where each layer consists of 500, 250, and 125 hidden nodes respectively.

Originally, 3D-CNN takes an input of $24 \times 10 \times 64 \times 64$ pixels, 
that is 24 data cubes that were captured over 24 different days across the entire growing season, where each data cube has a width and a height of 64 pixels, and 10 channels or feature rasters. 
In contrast, in this work we captured data only at the early stage of the growing season; therefore, we adapt 3D-CNN to process inputs with only one data cube of $5 \times 5 \times 8$ pixels.
Similarly, the original CNN-LF takes as input a data cube of $21 \times 21 \times 5$ pixels, so we adapt its architecture to accept our input dimensions. 
Note that AdaBoost.App, SAE, and MLR take in one-dimensional inputs (i.e., $1 \times 8$ pixels) while 3D-CNN and CNN-LF take in two-dimensional inputs (i.e., $5 \times 5 \times 8$ pixels); nevertheless, all these methods output a single yield value, unlike our Hyper3DNetReg architecture.   

For the sake of consistency and fairness, the six methods were trained using the same configuration; that is, the same loss function, optimizer and mini-batch size. 
All models were trained using mean squared error (MSE) as the loss function.
The selected optimizer was Adadelta~\cite{adadelta}, which is a gradient descent method based on an adaptive learning rate so that there is no need to select a global learning rate manually. 
The mini-batch size for all the fields was set to 96.
Note that min-max normalization was applied to each feature in the training set while the exact same scaling was applied to the features in the validation and test sets.

For each of the six yield prediction methods, we generated the predicted yield maps using the process described in Sec.~\ref{generate}.
To analyze the effectiveness of the methods, we calculated three metrics to compare the ground-truth yield map ($M$) and the predicted yield map ($\hat{M}$): Root mean square error ($RMSE$), root median square error ($RMedSE$), and structural similarity ($SSIM$). $RMSE$ and $RMedSE$ are calculated as follows:
\begin{equation*}
    RMSE = \sqrt{\frac{\sum_{(i,j) \in F} (M(i, j) - \hat{M}(i, j))^2}{|F|}} , 
\end{equation*}
    
\begin{equation*}
    RMedSE = \sqrt{\underset{(i,j) \in F}{\texttt{median}}(M(i, j) - \hat{M}(i, j))^2},
\end{equation*}
    
\noindent where $F$ represents the set of spatial coordinates that lie within the boundaries of the field. 

\begin{figure}[t!]
\centering
\includegraphics[width = 12cm]{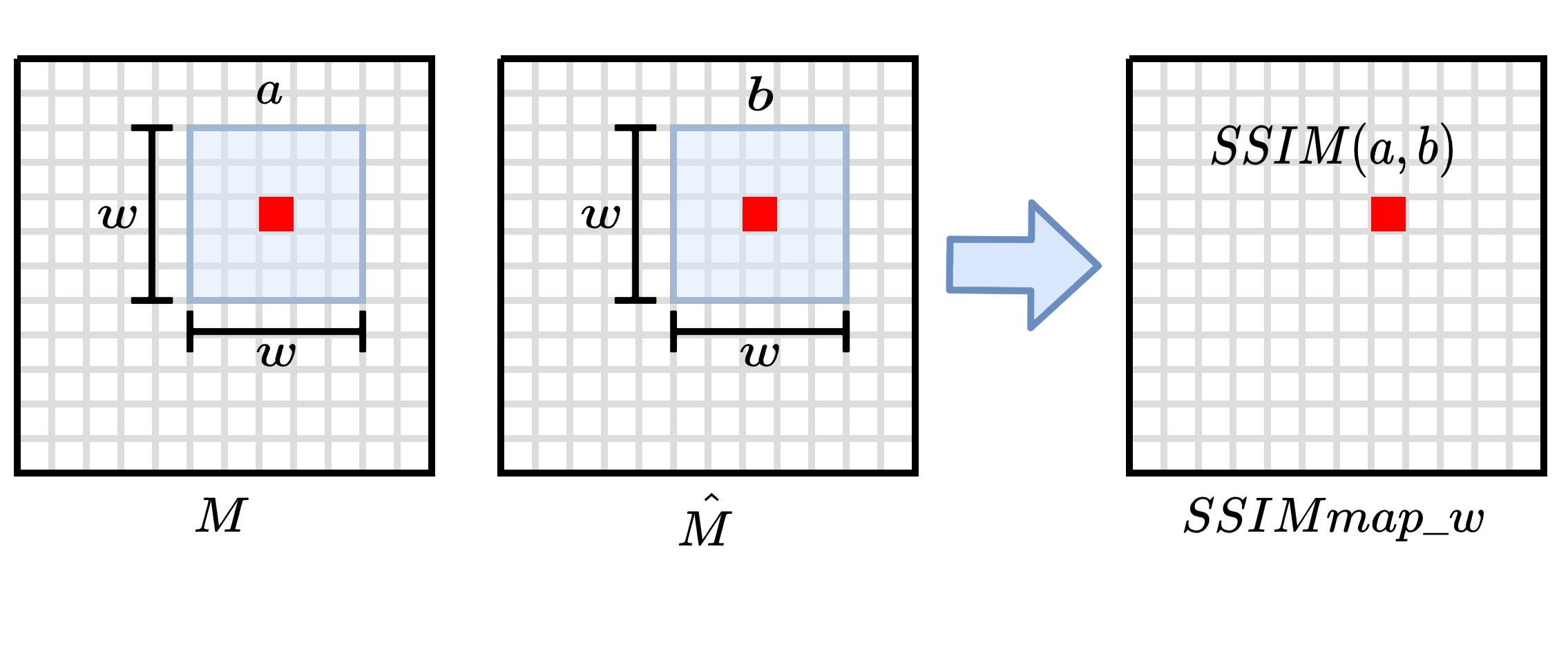}
\caption{Demonstration of the SSIM map generation between images $M$ and $\hat{M}$. $SSIM(a, b)$ represents the SSIM value between the local patches $a$ and $b$.} 
\label{fig:ssim}
\vspace{2ex}
\end{figure}

The SSIM index was originally proposed for image quality assessment to compare a noisy image with a raw image~\cite{ssim}. It combines three types of measurements: luminance, contrast, and structure. 
The resulting index is a single value between $-1$ and $+1$, where a value of $-1$ indicates no structural similarity, and a value of $+1$ indicates that both images are identical. 
Moreover, the SSIM index is calculated locally using $w \times w$ windows to create an SSIM map, as shown in Fig.~\ref{fig:ssim}.  
Even though our yield maps are not images \textit{per se}, we can treat them as such considering the yield values as pixel intensities. 

We generate two SSIM maps, $SSIM{map}\_3$ and $SSIM{map}\_11$, using two window sizes: $w=3$ and $w=11$, respectively. 
Smaller windows capture granular details while bigger windows focus on bigger structures.
Then, in order to obtain a single metric from the SSIM maps, we define the metrics $SSIM3$ and $SSIM11$ as follows:

\begin{equation*}
    SSIM3 = \frac{1}{|F|} \sum_{(i,j) \in F} SSIMmap\_3,
\end{equation*}
    
\begin{equation*}
    SSIM11 = \frac{1}{|F|} \sum_{(i,j) \in F} SSIMmap\_11.
\end{equation*}

\begin{figure}[t!]
\centering
\includegraphics[width = \textwidth]{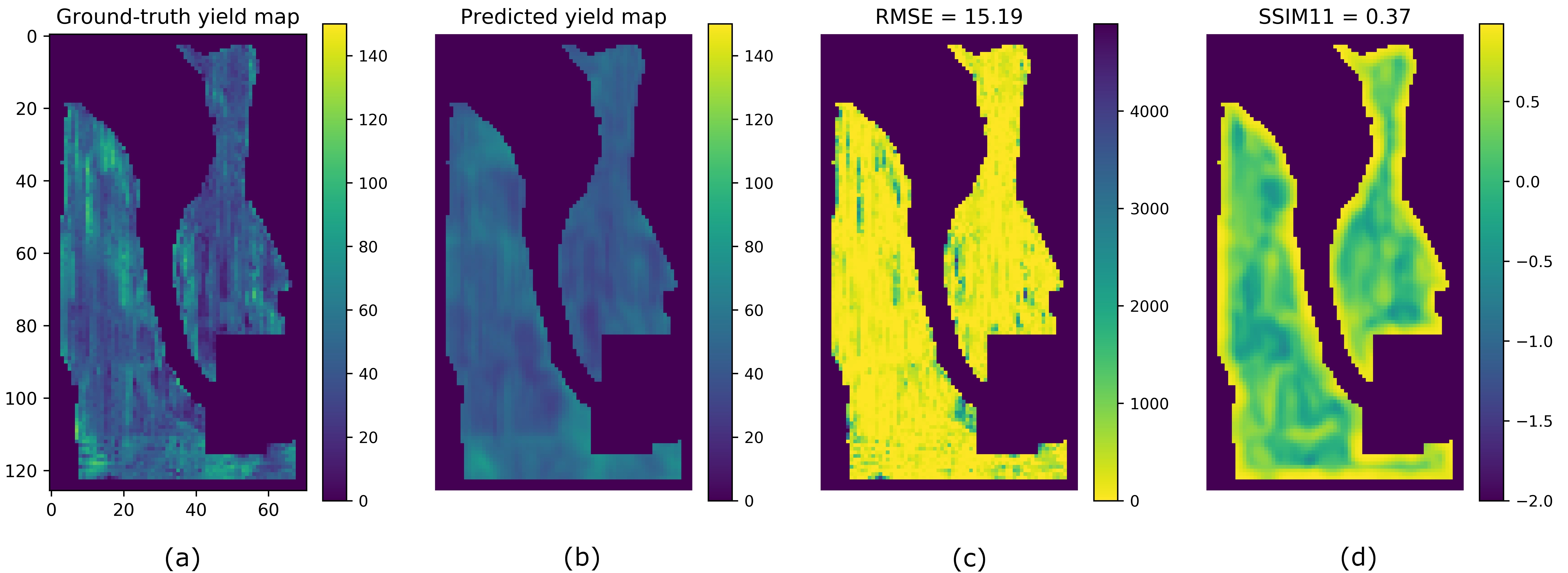}
\vspace{1ex}
\caption{Yield prediction example of the Sec35mid field. \textbf{(a)} Ground-truth yield map. \textbf{(b)} Predicted yield map using our Hyper3DNetReg with $N=5$. \textbf{(c)} Square error map. \textbf{(d)} $SSIMmap\_11$.} 
\label{fig:example}

\end{figure}

\setlength{\tabcolsep}{4pt}
\begin{table}
\caption{Yield prediction comparison.}
\label{tab:results}
\begin{center}
\def\arraystretch{1.2}%
\small
\begin{tabular}{|c|c|c|c|c|c|c|c|c|c|}
\hline
\textbf{Field} & \textbf{Metric} & \textbf{\begin{tabular}[c]{@{}c@{}}Hyper3D\\ NetReg\\ N = 1\end{tabular}} & \textbf{\begin{tabular}[c]{@{}c@{}}Hyper3D\\ NetReg\\ N = 3\end{tabular}} & \textbf{\begin{tabular}[c]{@{}c@{}}Hyper3D\\ NetReg\\ N = 5\end{tabular}} & \textbf{\begin{tabular}[c]{@{}c@{}}AdaBoost.\\ App\\ N = 1\end{tabular}} & \textbf{\begin{tabular}[c]{@{}c@{}}SAE\\ N=1\end{tabular}} & \textbf{\begin{tabular}[c]{@{}c@{}}3D-CNN\\ N=1\end{tabular}} & \textbf{\begin{tabular}[c]{@{}c@{}}CNN-LF\\ N=1\end{tabular}} & \textbf{\begin{tabular}[c]{@{}c@{}}MLR\\ N=1\end{tabular}} \\ \hline
 & $RMSE$ & 13.52 & 11.88 & 10.88 & 12.69 & 10.43 & 11.64 & \textbf{10.73} & 10.98 \\ \cline{2-10} 
 & $RMedSE$ & 8.94 & 8.10 & 7.01 & 7.74 & 6.75 & 7.59 & 7.42 & \textbf{6.74} \\ \cline{2-10} 
 & $SSIM3^*$ & 33.58 & 36.21 & \textbf{43.2} & 39.74 & 41.66 & 38.85 & 40.96 & 40.39 \\ \cline{2-10} 
\multirow{-4}{*}{Henrys} & $SSIM11^*$ & 51.87 & 56.29 & \textbf{62.23} & 58.82 & 60.99 & 57.81 & 60.15 & 59.8 \\ \hline
 & $RMSE$ & 17.05 & 21.22 & \textbf{16.71} & 17.02 & 37.33 & 24.52 & 21.1 & 28.09 \\ \cline{2-10} 
 & $RMedSE$ & 11.55 & 16.45 & \textbf{12.45} & 12.66 & 37.12 & 18.20 & 17.70 & 20.92 \\ \cline{2-10} 
 & $SSIM3^*$ & 5.44 & 5.52 & \textbf{6.55} & 5.92 & 1.73 & 3.3 & 5.56 & 1.88 \\ \cline{2-10} 
\multirow{-4}{*}{Sec1west} & $SSIM11^*$ & 13.96 & 12.69 & \textbf{15.38} & 14.43 & 5.59 & 8.9 & 13.51 & 3.35 \\ \hline
 & $RMSE$ & 17.39 & 15.93 & \textbf{15.19} & 16.05 & 15.66 & 18.94 & 16.81 & 16.85 \\ \cline{2-10} 
 & $RMedSE$ & 10.09 & 10.63 & \textbf{9.04} & 9.92 & 10.19 & 11.24 & 10.70 & 10.06 \\ \cline{2-10} 
 & $SSIM3^*$ & 15.93 & 15.44 & 15.95 & 19.73 & \textbf{22.21} & 22.35 & 23.15 & 24.51 \\ \cline{2-10} 
\multirow{-4}{*}{Sec35mid} & $SSIM11^*$ & 34.96 & 36.45 & 37.01 & 40.56 & \textbf{44} & 44.17 & 46.03 & 46.61 \\ \hline
 & $RMSE$ & 19.28 & 19.11 & \textbf{16.62} & 23.19 & 17.62 & 21.44 & 42.86 & 27.57 \\ \cline{2-10} 
 & $RMedSE$ & 14.51 & 13.75 & \textbf{11.26} & 16.71 & 13.32 & 13.73 & 14.88 & 19.19 \\ \cline{2-10} 
 & $SSIM3^*$ & 25.36 & 27.22 & \textbf{29.49} & 25.2 & 27.81 & 27.37 & 23.5 & 23.12 \\ \cline{2-10} 
\multirow{-4}{*}{Sec35west} & $SSIM11^*$ & 48.24 & 51.43 & \textbf{54.05} & 48.32 & 52.55 & 47.79 & 37.78 & 39.61 \\ \hline
\end{tabular}

\end{center}
\end{table}

Fig.~\ref{fig:example}.b illustrates the predicted yield map that results from applying our Hyper3DNetReg network with an output size of five pixels ($N=5$). 
Fig.~\ref{fig:example}.c shows the corresponding square error map calculated as $(M - \hat{M})^2$. 
Moreover, Fig.~\ref{fig:example}.d shows the SSIM map obtained using a window size of 11 pixels; that is, $SSIMmap\_11$. 
Note that $SSIMmap\_11$ presents high values in the regions with higher similarity and also in the borders. 
This is because pixels located near the field boundary generate a $11 \times 11$ window around them to calculate the SSIM index (as shown in Fig.~\ref{fig:ssim}). 
Therefore, these windows end up including information from the background (i.e. zero-valued pixels). 
Given that both windows extracted from $M$ and $\hat{M}$ contain background pixels at the same locations and with identical values, the resulting SSIM index is high.

\begin{figure}[t!]
\centering
\includegraphics[width = 16cm]{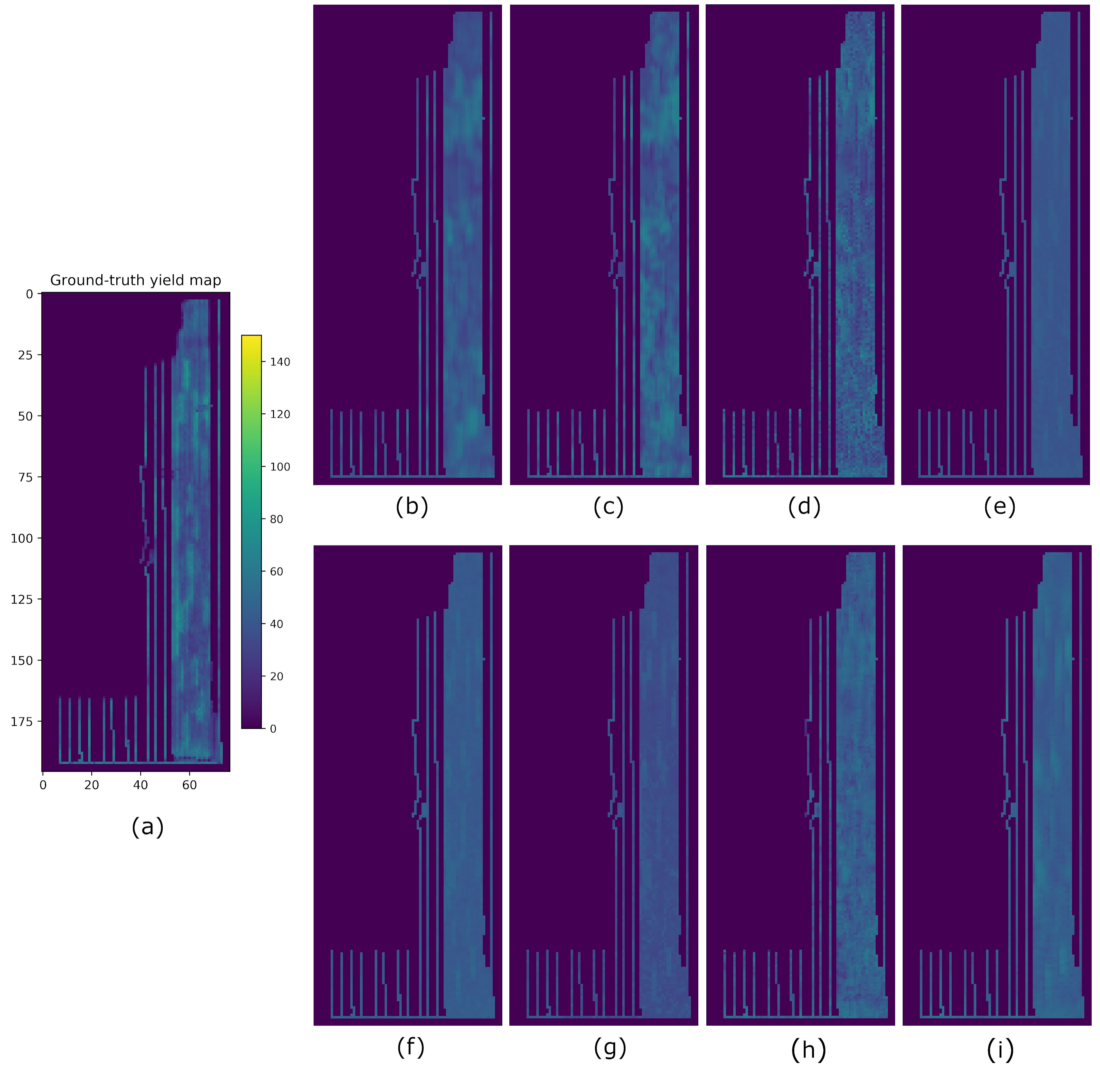}
\caption{Yield prediction comparison on the Henrys field. \textbf{(a)} Ground-truth. \textbf{(b)} Hyper3DNetReg ($N=5$). \textbf{(c)} Hyper3DNetReg ($N=3$). \textbf{(d)} Hyper3DNetReg ($N=1$). \textbf{(e)} MLR. \textbf{(f)} SAE. \textbf{(g)} AdaBoost. \textbf{(h)} 3D-CNN. \textbf{(i)} CNN-LF.} 
\label{fig:henrys}
\end{figure}

\begin{figure}[t!]
\centering
\includegraphics[width = 16cm]{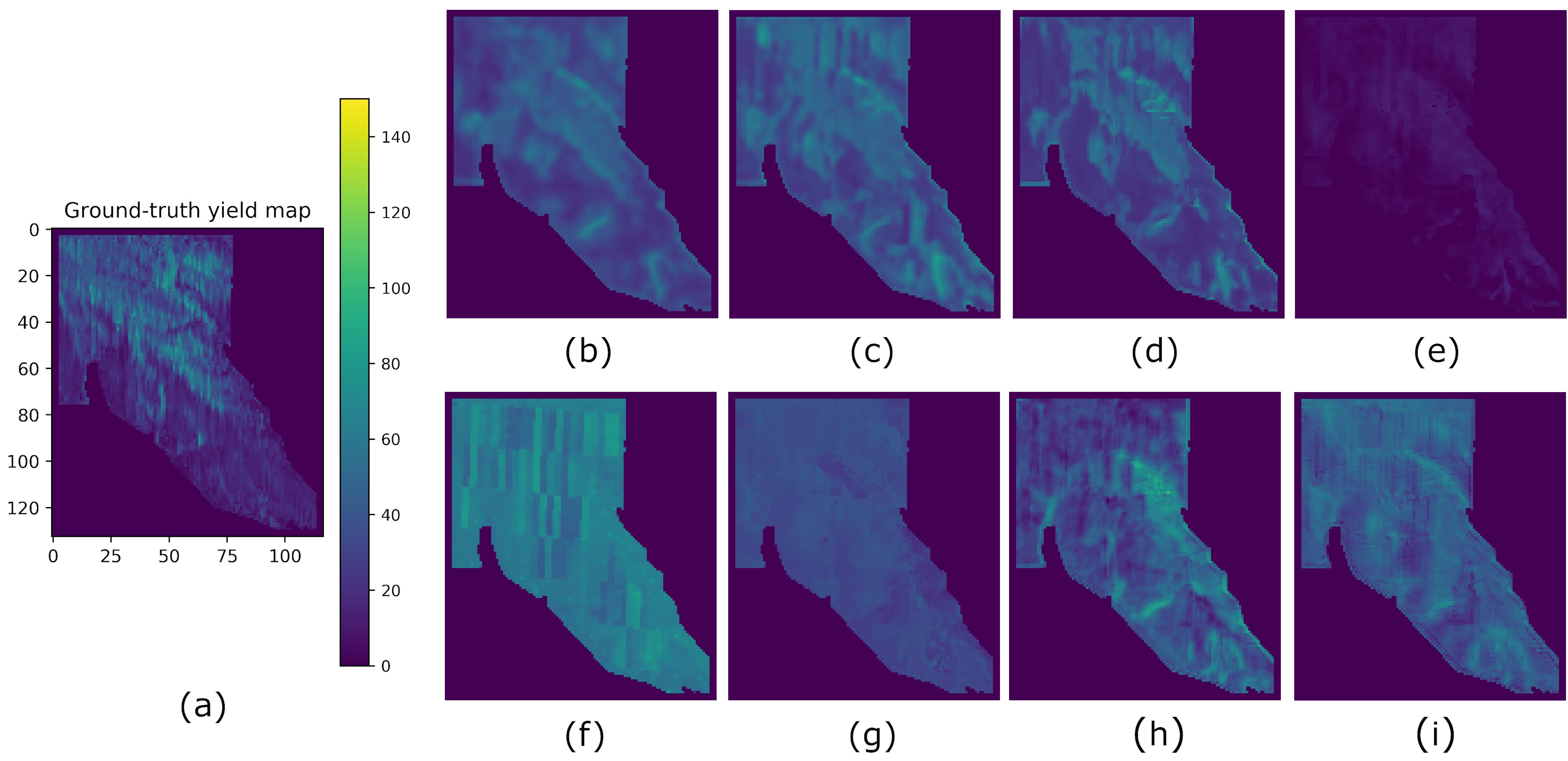}
\caption{Yield prediction comparison on the Sec1west field. \textbf{(a)} Ground-truth. \textbf{(b)} Hyper3DNetReg ($N=5$). \textbf{(c)} Hyper3DNetReg ($N=3$). \textbf{(d)} Hyper3DNetReg ($N=1$). \textbf{(e)} MLR. \textbf{(f)} SAE. \textbf{(g)} AdaBoost. \textbf{(h)} 3D-CNN. \textbf{(i)} CNN-LF.} 
\label{fig:sec1w}
\end{figure}

\begin{figure}[t!]
\centering
\includegraphics[width = 16cm]{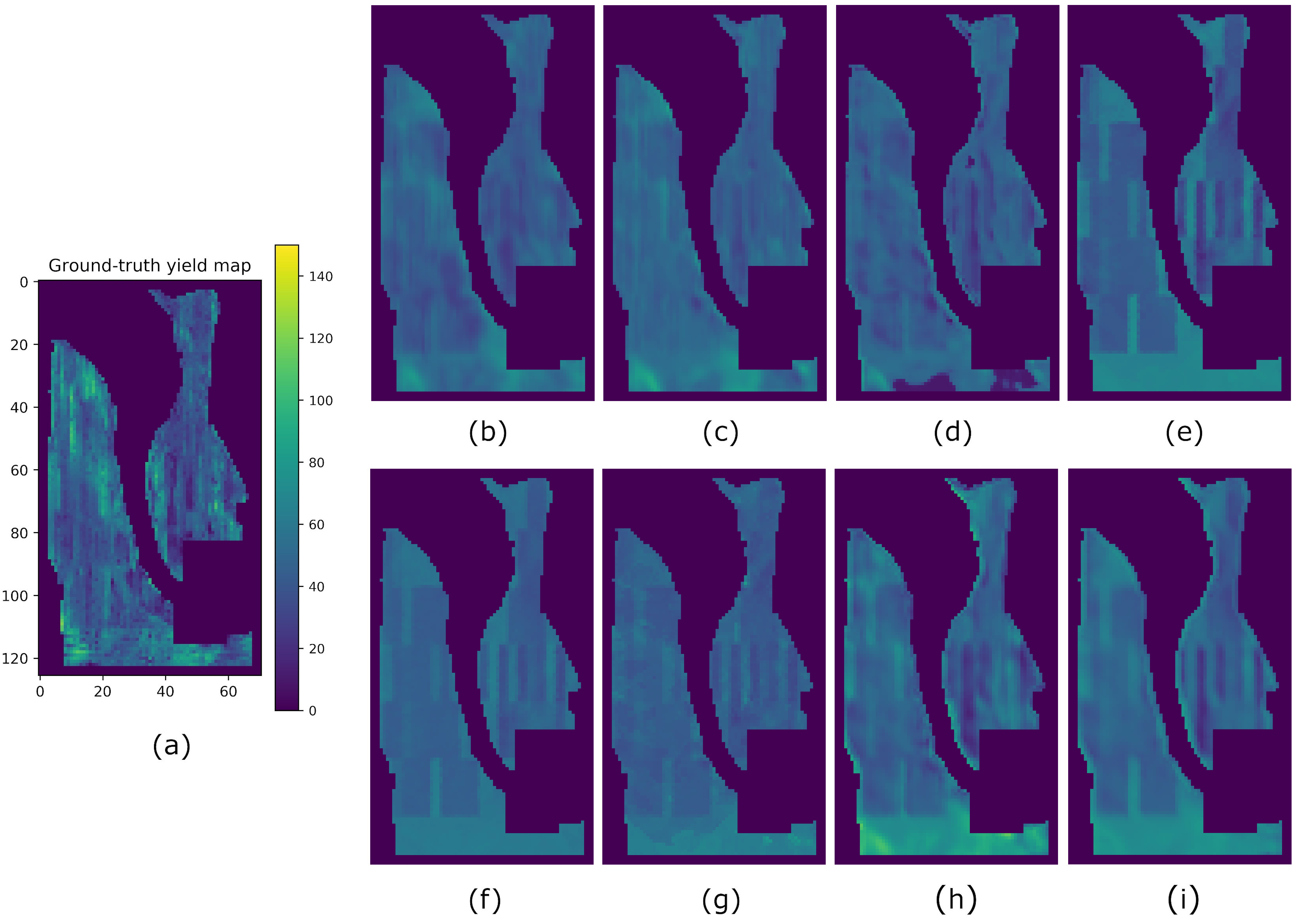}
\caption{Yield prediction comparison on the Sec35mid field. \textbf{(a)} Ground-truth. \textbf{(b)} Hyper3DNetReg ($N=5$). \textbf{(c)} Hyper3DNetReg ($N=3$). \textbf{(d)} Hyper3DNetReg ($N=1$). \textbf{(e)} MLR. \textbf{(f)} SAE. \textbf{(g)} AdaBoost. \textbf{(h)} 3D-CNN. \textbf{(i)} CNN-LF.} 
\label{fig:sec35m}
\end{figure}

\begin{figure}[t!]
\centering
\includegraphics[width = 16cm]{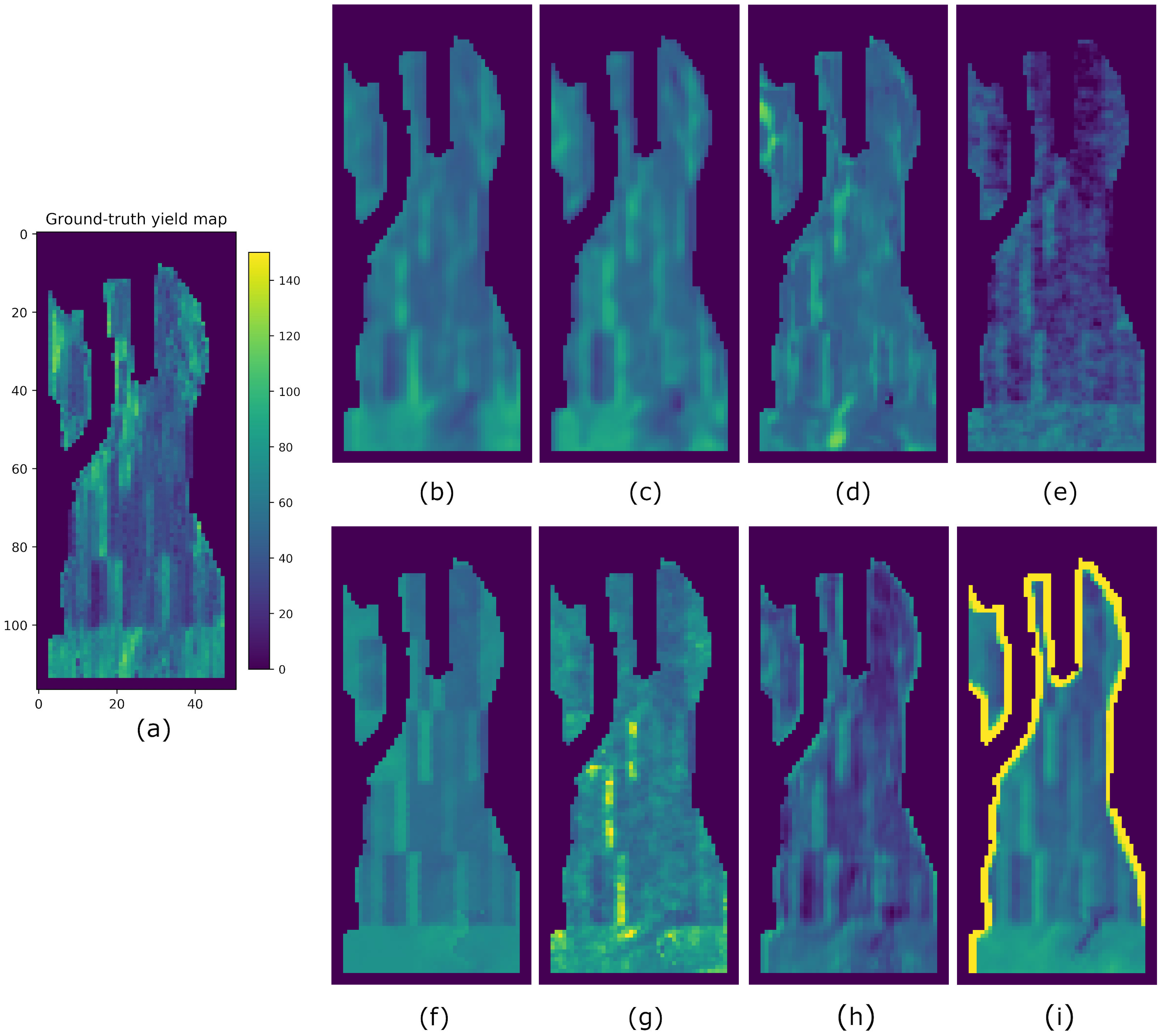}
\caption{Yield prediction comparison on the Sec35west field. \textbf{(a)} Ground-truth. \textbf{(b)} Hyper3DNetReg ($N=5$). \textbf{(c)} Hyper3DNetReg ($N=3$). \textbf{(d)} Hyper3DNetReg ($N=1$). \textbf{(e)} MLR. \textbf{(f)} SAE. \textbf{(g)} AdaBoost. \textbf{(h)} 3D-CNN. \textbf{(i)} CNN-LF.} 
\label{fig:sec35w}
\end{figure}

The method comparison is shown in Table~\ref{tab:results}.
Here, the bold entries represent the best metrics for each field.
$SSIM3^*$ and $SSIM11^*$ indicate that the original SSIM indices have been multiplied by 100 for easier visualization and comparison of the results.
Finally, Fig.~\ref{fig:henrys}, ~\ref{fig:sec1w}, ~\ref{fig:sec35m}, and ~\ref{fig:sec35w} depict the predicted yield maps of different methods for the Henrys, Sec1west, Sec35mid, and Sec35west fields, respectively.

\section{Discussion} \label{sec:discussion}

From Table ~\ref{tab:results}, we conclude that our Hyper3DNetReg achieved the best results on the fields studied when using an output window size of five pixels.
We denote this configuration as Hyper3DNetReg-N5 for conciseness. 
For the case of the Henrys field, SAE, CNN-LF, and MLR achieved RMSE values of 10.43, 10.73 and 10.98, respectively, which are comparable to that obtained by Hyper3DNetReg-N5, 10.88.
However, we are not only interested in reducing the global $RMSE$.
We would like that the high and low yield regions of our predicted yield maps correspond to actual high and low regions, which can be verified by measuring the structural similarity between the predicted yield map and the ground-truth yield map locally.
Therefore, we need to interpret the $RMSE$ metric alongside the $SSIM3$ and $SSIM11$ metrics.
By doing so, we notice that Hyper3DNetReg-N5 yields one of the lowest $RMSE$ but also the highest $SSIM3$ and $SSIM11$ values, which is why it is considered the best yield prediction method for this field.
For the rest of the fields, Hyper3DNetReg-N5 achieved the best metrics, except for Sec35mid, where the other methods present higher $SSIM3$ and $SSIM11$ values. We will discuss this issue below.

\begin{figure}[t!]
\centering
\includegraphics[width = 12cm]{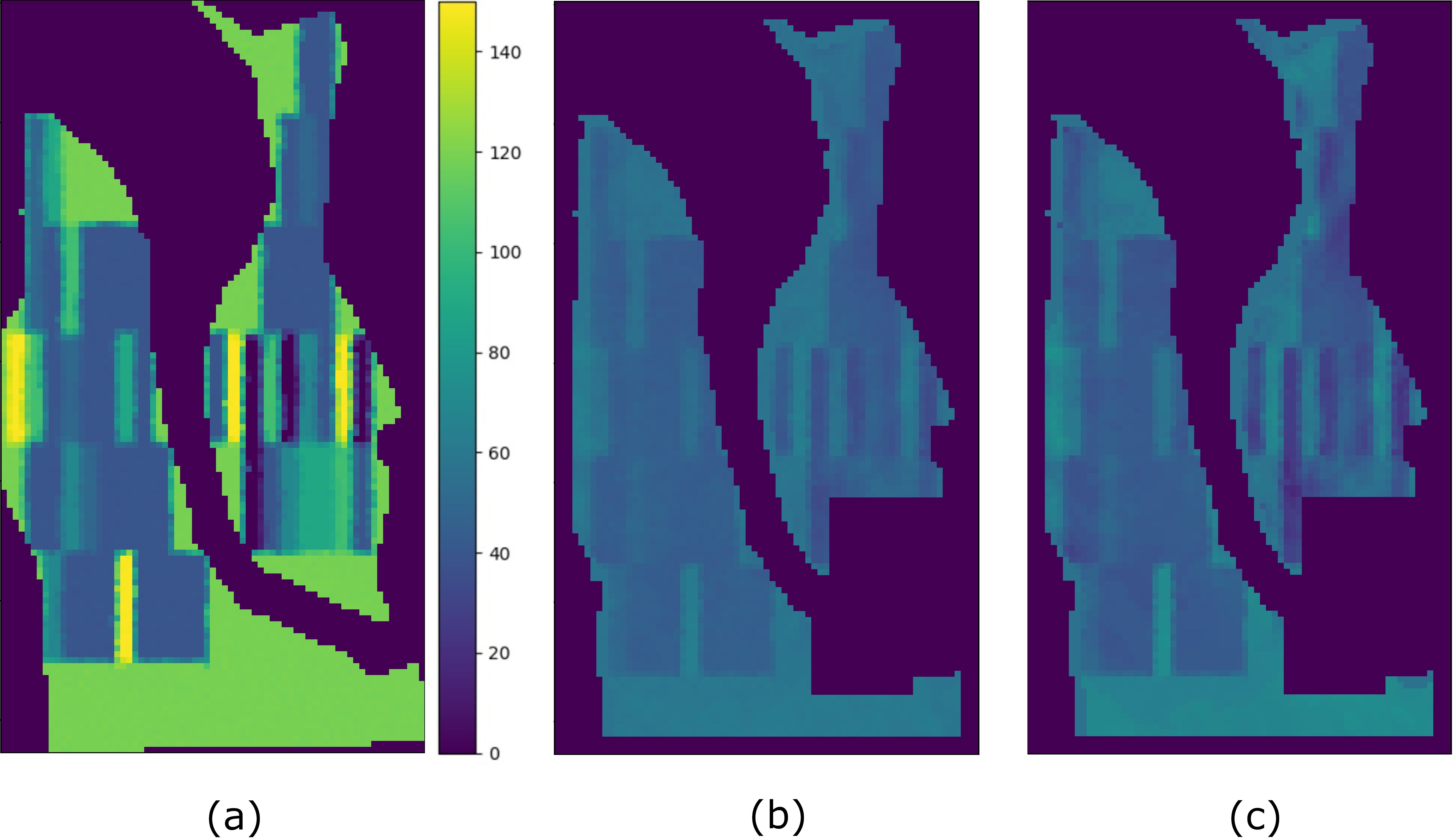}
\caption{Comparison of the nitrogen rate map and the predicted yield map on the Sec35middle field. \textbf{(a)} Nitrogen rate map. \textbf{(b)} SAE. \textbf{(c)} MLR.} 
\label{fig:nvsy}
\end{figure}

From visual inspection of the results presented in Fig.~\ref{fig:henrys}--\ref{fig:sec35w}, we can conclude that our method predicts yield maps that are more similar to the ground truth than the yield maps predicted by the other methods. 
Continuing our discussion of the results obtained on the Henrys field, Hyper3DNetReg-N5 is the only method that succeeds at identifying regions of high and low yield, as seen in Fig.~\ref{fig:henrys}.b, which led it to achieve the highest $SSIM3$ and $SSIM11$ values. 
However, as we mentioned above, other methods, such as SAE, present higher SSIM values than Hyper3DNetReg-N5 for the Sec35mid field.
This is because the yield map generated by SAE (Fig.~\ref{fig:sec35m}.f) presents some sharp regions of high yield that coincide with the ground-truth yield map and increase the overall SSIM metrics.
The yield map predicted by Hyper3DNetReg-N5 also detects those high yield regions but the resulting structural similarity is lower because of the blur that follows from averaging overlapping regions; thus, it obtains lower SSIM values despite better representing the ground-truth.
It is worth mentioning that the yield map predicted by SAE, as well as that predicted by MLR (Fig.~\ref{fig:sec35m}.e), shows that behavior because it considers the nitrogen rate as the most important explanatory variable.
This results in yield prediction maps that are structurally similar to the nitrogen rate maps, as shown in Fig.~\ref{fig:nvsy}. 
Future work will focus on measuring the relevance each variable has when predicting yield values.

The Sec35west field is a special case because it has fewer training samples than the rest of the fields and only one year of training data.
Furthermore, the testing year contains information that was not seen in the training year.
For instance, the average precipitation value observed in the training year was \SI{109}{\nano\meter} while the average precipitation value observed in the testing year was \SI{67}{\nano\meter}.
In addition, the training set only included patch samples from the inner regions of the field and not from the regions next to the boundary because they had missing nitrogen and yield information.
Thus, the limited amount of training data and its lack of diversity cause the models to be prone to overfitting.
This can be seen with CNN-LF (see Fig.~\ref{fig:sec35w}.i)  
where the overfitting arises due to the fact CNN-LF is a multi-stream network that analyzes each input variable separately and fuses the outputs from each stream using a single neuron.
The weight this final neuron assigned to the precipitation stream is negative and large in magnitude.
Hence, the predicted yield is higher when it receives lower precipitation values than what the network used for training.
In this case, the precipitation values outside the boundary of the field are equal to zero, which causes high predicted yield values in these regions.
It is worth noting that the training set used for Sec35west only included patch samples from the inner regions of the field and not from the regions next to the boundary because they had missing nitrogen and yield information.

In general, Hyper3DNetReg-N5 achieved better predictions than the compared methods that use a single output, which confirms our hypothesis. 
What is more, we observe in Table~\ref{tab:results} that Hyper3DNetReg-N5 consistently yields better results than Hyper3DNetReg when it uses output sizes of three pixels and one pixel (Hyper3DNetReg-N3 and Hyper3DNetReg-N1, respectively). 
In addition, Hyper3DNetReg-N3 consistently outperforms Hyper3DNetReg-N1. 
Take, for example, Fig.~\ref{fig:sec35w}; here, we can see that the map generated by Hyper3DNetReg-N5 (Fig.~\ref{fig:sec35w}.b) is less noisy than that generated by Hyper3DNetReg-N3 (Fig.~\ref{fig:sec35w}.c), which, in turn, is less noisy than that generated by Hyper3DNetReg-N1 (Fig.~\ref{fig:sec35w}.d). 
This can be explained by the fact that the bigger the output window size, the more the number of pixels that overlap during the generation of the predicted yield map (as explained in Sec.~\ref{generate}). Therefore, we obtain multiple measurements of the estimated yield at each point of the field and then take the average value. This helps to reduce uncertainty and the effect of outlier responses.

\section{Conclusions and Future Work} \label{sec:conclusions}

Early crop yield prediction allows farmers to better allocate farming resources. 
For instance, the timely identification of a region of the field that will present a low yield during the harvest season allows farmers to take actions that compensate for this behavior, such as increasing the nitrogen rate.
Therefore, it is crucial to develop accurate yield prediction models to take full advantage of the available technology and maximize farmer profits.

In this paper, we presented a convolutional neural network architecture called Hyper3DNetReg.
Our network tackles the yield prediction problem as a two-dimensional regression task 
and allows predicting the yield values of a small spatial neighborhood of a field simultaneously.
Experimental results show that our model generates the most accurate predicted yield maps in comparison to other models that use a single output.

Future work will focus on the use of our models for the detection of nitrogen-responsive regions within the field through a saliency analysis. In addition, we will modify our models to calculate narrow and accurate prediction intervals automatically so as to offer a degree of confidence alongside each prediction.

\acknowledgments{The authors wish to thank the team members of the On-Field Precision Experiment (OFPE) project for their comments through the development of this work, especially Paul Hegedus for collecting and curating the site-specific data, and Amy Peerlinck for providing the Approximation AdaBoost code and assistance with its utilization. This research was supported by a USDA-NIFA-AFRI Food Security Program Coordinated Agricultural Project, titled “Using Precision Technology in On-farm Field Trials to Enable Data-Intensive Fertilizer Management,” (Accession Number 2016-68004-24769), and also by the USDA-NRCS Conservation Innovation Grant from the On-farm Trials Program, titled “Improving the Economic and Ecological Sustainability of US Crop Production through On-Farm Precision Experimentation” (Award Number NR213A7500013G021).}

\bibliography{report} 

\begin{thebibliography}{10}

\bibitem{future}
McBratney, A., Whelan, B., Ancev, T., and Bouma, J., ``Future directions of
  precision agriculture,'' {\em Precision Agriculture}~{\bf 6},  7--23 (2005).

\bibitem{PAapplications}
Shafi, U., Mumtaz, R., García-Nieto, J., Hassan, S., Zaidi, S., and Iqbal, N.,
  ``Precision agriculture techniques and practices: From considerations to
  applications,'' {\em Sensors}~{\bf 19}(17) (2019).

\bibitem{PAreview}
Cisternas, I., Velásquez, I., Caro, A., and Rodríguez, A., ``Systematic
  literature review of implementations of precision agriculture,'' {\em
  Computers and Electronics in Agriculture}~{\bf 176},  105626 (2020).

\bibitem{bullock}
Bullock, D.~S., Boerngen, M., Tao, H., Maxwell, B., Luck, J.~D., Shiratsuchi,
  L., Puntel, L., and Martin, N.~F., ``The data-intensive farm management
  project: Changing agronomic research through on-farm precision
  experimentation,'' {\em Agronomy Journal}~{\bf 111}(6),  2736--2746 (2019).

\bibitem{IOUT}
Vuran, M., Salam, A., Wong, R., and Irmak, S., ``Internet of underground things
  in precision agriculture: Architecture and technology aspects,'' {\em Ad Hoc
  Networks}~{\bf 81},  160 -- 173 (2018).

\bibitem{UAS}
Hunt, E. and Daughtry, C. S.~T., ``What good are unmanned aircraft systems for
  agricultural remote sensing and precision agriculture?,'' {\em International
  Journal of Remote Sensing}~{\bf 39}(15-16),  5345--5376 (2018).

\bibitem{yieldML}
{Van Klompenburg}, T., Kassahun, A., and Catal, C., ``Crop yield prediction
  using machine learning: A systematic literature review,'' {\em Computers and
  Electronics in Agriculture}~{\bf 177},  105709 (2020).

\bibitem{CVreview}
Patrício, D. and Rieder, R., ``Computer vision and artificial intelligence in
  precision agriculture for grain crops: A systematic review,'' {\em Computers
  and Electronics in Agriculture}~{\bf 153},  69 -- 81 (2018).

\bibitem{HORIE}
Horie, T., Yajima, M., and Nakagawa, H., ``Yield forecasting,'' {\em
  Agricultural Systems}~{\bf 40}(1),  211--236 (1992).

\bibitem{maxwell}
Maxwell, B., Hegedus, P., Davis, P., Bekkerman, A., Payn, R., Sheppard, J.,
  Silverman, N., and Izurieta, C., ``Can optimization associated with on-farm
  experimentation using site-specific technologies improve producer management
  decisions?,'' in [{\em Proceedings of the 14th International Conference on
  Precision Agriculture}{\nolinebreak\hspace{0.1em}]},  (2018).

\bibitem{msthesis}
{Russello}, H., {\em Convolutional Neural Networks for Crop Yield Prediction
  using Satellite Images}, Master's thesis, University of Amsterdam (2018).

\bibitem{potato}
Gómez, D., Salvador, P., Sanz, J., and Casanova, J., ``Potato yield prediction
  using machine learning techniques and {Sentinel} 2 data,'' {\em Remote
  Sensing}~{\bf 11}(15) (2019).

\bibitem{barbosa}
Barbosa, A., Trevisan, R., Hovakimyan, N., and Martin, N., ``Modeling yield
  response to crop management using convolutional neural networks,'' {\em
  Computers and Electronics in Agriculture}~{\bf 170},  105197 (2020).

\bibitem{peerlinck2019adaboost}
Peerlinck, A., Sheppard, J., and Senecal, J., ``Adaboost with neural networks
  for yield and protein prediction in precision agriculture,'' in [{\em
  Proceedings of the International Joint Conference on Neural Networks
  (IJCNN)}{\nolinebreak\hspace{0.1em}]},  IEEE (2019).

\bibitem{reg1}
Bolton, D. and Friedl, M., ``Forecasting crop yield using remotely sensed
  vegetation indices and crop phenology metrics,'' {\em Agricultural and Forest
  Meteorology}~{\bf 173},  74--84 (2013).

\bibitem{reg2}
Johnson, D., ``An assessment of pre- and within-season remotely sensed
  variables for forecasting corn and soybean yields in the united states,''
  {\em Remote Sensing of Environment}~{\bf 141},  116--128 (2014).

\bibitem{svm}
Kim, N. and Lee, Y., ``Machine learning approaches to corn yield estimation
  using satellite images and climate data: A case of iowa state,'' {\em Journal
  of the Korean Society of Surveying, Geodesy, Photogrammetry and
  Cartography}~{\bf 34},  383--390 (08 2016).

\bibitem{rf}
Wei, M., Maldaner, L., Ottoni, P., and Molin, J., ``Carrot yield mapping: A
  precision agriculture approach based on machine learning,'' {\em AI}~{\bf
  1}(2),  229--241 (2020).

\bibitem{regtree}
Gonzalez-Sanchez, A., Frausto-Solis, J., and Ojeda-Bustamante, W., ``Predictive
  ability of machine learning methods for massive crop yield prediction,'' {\em
  Spanish Journal of Agricultural Research}~{\bf 12}(2) (2014).

\bibitem{Peerlinck2018UsingDL}
Peerlinck, A., Sheppard, J., and Maxwell, B., ``Using deep learning in yield
  and protein prediction of winter wheat based on fertilization prescriptions
  in precision agriculture,'' in [{\em Proceedings of the 14th International
  Conference on Precision Agriculture}{\nolinebreak\hspace{0.1em}]},  (2018).

\bibitem{deepgauss}
You, J., Li, X., Low, M., Lobell, D., and Ermon, S., ``Deep gaussian process
  for crop yield prediction based on remote sensing data,'' in [{\em
  Proceedings of the AAAI Conference on Artificial
  Intelligence}{\nolinebreak\hspace{0.1em}]},  (2017).

\bibitem{moisture1}
Álvarez Mozos, J., Verhoest, N.~E., Larrañaga, A., Casalí, J., and
  González-Audícana, M., ``Influence of surface roughness spatial variability
  and temporal dynamics on the retrieval of soil moisture from {SAR}
  observations,'' {\em Sensors}~{\bf 9}(1),  463--489 (2009).

\bibitem{moisture2}
Zhang, L., Lv, X., Chen, Q., Sun, G., and Yao, J., ``Estimation of surface soil
  moisture during corn growth stage from {SAR} and optical data using a
  combined scattering model,'' {\em Remote Sensing}~{\bf 12}(11) (2020).

\bibitem{vegetation}
Vreugdenhil, M., Wagner, W., Bauer-Marschallinger, B., Pfeil, I., Teubner, I.,
  Rüdiger, C., and Strauss, P., ``Sensitivity of sentinel-1 backscatter to
  vegetation dynamics: An austrian case study,'' {\em Remote Sensing}~{\bf
  10}(9) (2018).

\bibitem{biomass}
Betbeder, J., Fieuzal, R., Philippets, Y., Ferro-Famil, L., and Baup, F.,
  ``{Contribution of multitemporal polarimetric synthetic aperture radar data
  for monitoring winter wheat and rapeseed crops},'' {\em Journal of Applied
  Remote Sensing}~{\bf 10}(2),  1 -- 19 (2016).

\bibitem{rice}
Clauss, K., Ottinger, M., Leinenkugel, P., and Kuenzer, C., ``Estimating rice
  production in the {Mekong Delta, Vietnam}, utilizing time series of
  {Sentinel-1} {SAR} data,'' {\em International Journal of Applied Earth
  Observation and Geoinformation}~{\bf 73},  574--585 (2018).

\bibitem{moistureyield}
Zhuo, W., Huang, J., Li, L., Zhang, X., Ma, H., Gao, X., Huang, H., Xu, B., and
  Xiao, X., ``Assimilating soil moisture retrieved from {Sentinel-1} and
  {Sentinel-2} data into {WOFOST} model to improve winter wheat yield
  estimation,'' {\em Remote Sensing}~{\bf 11}(13) (2019).

\bibitem{van1989wofost}
Van~Diepen, C., Wolf, J., Van~Keulen, H., and Rappoldt, C., ``{WOFOST: A
  simulation model of crop production},'' {\em Soil Use and Management}~{\bf
  5}(1),  16--24 (1989).

\bibitem{hyper3dnet}
{Morales}, G., {Sheppard}, J., {Scherrer}, B., and {Shaw}, J., ``Reduced-cost
  hyperspectral convolutional neural networks,'' {\em Journal of Applied Remote
  Sensing}~{\bf 14}(3),  036519 (2020).

\bibitem{adadelta}
Zeiler, M.~D., ``{ADADELTA}: An adaptive learning rate method,'' (2012).
\newblock arXiv:1212.5701.

\bibitem{ssim}
Wang, Z., Bovik, A., Sheikh, H., and Simoncelli, E., ``{Image quality
  assessment: From error visibility to structural similarity},'' {\em IEEE
  Transactions on Image Processing}~{\bf 13}(4),  600--612 (2004).

\end{thebibliography}
\bibliographystyle{spiebib} 

\end{document}